%% file: TKDE.tex
\documentclass{article}
\usepackage{algorithm}
\usepackage{algorithmic}
\usepackage{graphicx}
\usepackage{amsthm}
\usepackage{amsmath}
\usepackage{bbm}
\title{Differential Privacy and Machine Learning: 
\\a Survey and Review}
\begin{document}
\author{Zhanglong~Ji, Zachary~C.~Lipton, Charles~Elkan}
\maketitle

\begin{abstract}
The objective of machine learning is to extract useful information from data, while privacy is preserved by concealing information. Thus it seems hard to reconcile these competing interests. However, they frequently must be balanced when mining sensitive data. For example, medical research represents an important application where it is necessary both to extract useful information and protect patient privacy. One way to resolve the conflict is to extract general characteristics of whole populations without disclosing the private information of individuals. 

In this paper, we consider differential privacy, one of the most popular and powerful definitions of privacy. 
We explore the interplay between machine learning and differential privacy, 
namely privacy-preserving machine learning algorithms 
and learning-based data release mechanisms. 
We also describe some theoretical results that address what can be learned differentially privately and upper bounds of loss functions for differentially private algorithms. 

Finally, we present some open questions, 
including how to incorporate public data, 
how to deal with missing data in private datasets, 
and whether, as the number of observed samples grows arbitrarily large,
differentially private machine learning algorithms can be achieved at no cost to utility as compared to corresponding non-differentially private algorithms. 
\end{abstract}

The objective of machine learning is to extract useful information from data, such as how to classify data, how to predict a quantity, or how to find clusters of similar samples. Given a family of learning models, machine learning algorithms select and output the best one based on some given data. The output model can be used in either dealing with future data or interpreting the distribution of data. 
Although the output model typically far more compact than the underlying dataset, it must capture some information describing the dataset. Privacy, on the other hand, concerns the protection of private data from leakage, especially the information of individuals.\footnote{
\cite{DBLP:dblp_conf/kdd/Cormode11} uses \textit{privacy} to mean both \emph{population privacy} and \emph{individual privacy}. For example, disclosing that members of some family are highly susceptible to a given genetic condition might violate population privacy, while disclosing that some specific patient suffers from this condition would violate their \emph{individual privacy}. However, even if one conceals all his/her information, it's still possible to breach population privacy by collecting information from other members of the subpopulation. As one cannot easily protect oneself from the disclosure of this information, we will use \textit{privacy} to refer only to individual privacy in this paper.}

It would be reasonable to ask,  ``why is it insufficient to anonymize data?" 
One could remove names and other obviously identifiable information from a database.
It might seem difficult then, for an attacker to identify an individual. Current HIPAA guidelines promotes such an approach, listing 18 categories of personally identifiable information which must be redacted in the case of research data for publication. Unfortunately, this method can leak information when the attacker already has some information about the individuals in question. In a well-known case, the personal health information of Massachusetts governor William Weld was discovered in a supposedly anonymized public database \cite{DBLP:dblp_journals/ijufks/Sweene02}. By merging overlapping records between the health database and a voter registry, researchers were able to identify the personal health records of the governor, among others.

To combat such background attacks, some more robust definitions of privacy (such as $k$-anonymity\cite{DBLP:dblp_journals/ijufks/Sweene02}, $l$-diversity\cite{DBLP:dblp_conf/icde/MachanavajjhalaGKV06} and $t$-closeness\cite{DBLP:dblp_conf/icde/LiLV07}) have been proposed. In these approaches, samples are grouped if their sensitive features are the same, and a group is published if the number of samples in that group is large enough. Intuitively, it should be difficult for an attacker to distinguish individual samples. However, even these definitions cannot prevent background attacks, in which the attackers already know something about the information contained in the dataset.  In an extreme case, the attacker might know the contents of all but one of the rows in the set. 

Consider a database which holds the address and income of four people 
and publishes private data using 3-anonymity. 
According to 3-anonymity, if any three people live in the same city, 
the city and average income of the three people is released. 
Now suppose an attacker knows that two people in the database live in Los Angeles 
and that a third lives in New York. 
If no data is published, the attacker can easily infer 
that the fourth person does not live in Los Angeles.

Even if one released aggregated statistics, they might risk compromising private information. 
Recently \cite{snps}, researchers demonstrated that an attacker could infer 
whether an individual had participated in a genome study 
using only publicly available aggregated genetic data. 
In such cases, aggregation is no longer safe. 

\emph{Differential privacy} \cite{DBLP:dblp_reference/crypt/Dwork11, DBLP:dblp_conf/eurocrypt/DworkKMMN06}, 
which will be introduced in the next section, 
uses random noise to ensure that the publicly visible information 
doesn't change much if one individual in the dataset changes. 

As no individual sample can significantly affect the output, 
attackers cannot infer the private information corresponding 
to any individual sample confidently. 
This paper addresses the interplay between machine learning 
and differential privacy.

Although it seems that machine learning and privacy protection are in opposition,
it is often possible to reconcile them. 
Researchers have designed many mechanisms to build models 
that can capture the distributions corresponding to large datasets 
while guaranteeing differential privacy with respect to individual examples. 
In order to achieve generalizability, machine learning models 
should not depend heavily on any single sample.
Therefore, it is possible to hide the effects of individual samples,
simultaneously preserving privacy and providing utility.

\subsection{Prior Work}
Several recent surveys address differential privacy and data science \cite{survey1, DBLP:dblp_journals/spm/SarwateC13, survey3}.
Some others (\cite{survey1, survey3}) mainly focus on statistical estimators, 
while \cite{DBLP:dblp_journals/spm/SarwateC13} discusses the high level interactions between differential privacy and machine learning.

Our survey focuses specifically on methods by which machine learning algorithms can be made differentially private. 
We study current differentially private machine learning algorithms 
and organize them according to the fundamental machine learning tasks they address, including classification, regression, clustering, and dimensionality reduction. 
We also describe some differentially private data release mechanisms, 
both because their mechanisms involves differential privacy, 
and because their output can be used to learn differentially private machine learning models. 
We explain how all of these mechanisms work 
and compare their theoretical guarantees. 
Some general theoretical results and discussion follow.

\section{Differential Privacy}
Differential privacy is one of the most popular definitions of privacy today. 
Intuitively, it requires that the mechanism outputting information 
about an underlying dataset is robust to any change of one sample, thus protecting privacy.

The following subsections mathematically define differential privacy and introduce some commonly used methods in differential privacy.

\subsection{Definition of Differential Privacy}
\textbf{Definition 1:} a \emph{mechanism} $\tilde{f}$ is a random function that takes a dataset $D$ as input, and outputs a random variable $\tilde{f}(D)$. 

For example, suppose $D$ is a medical dataset, then the function that outputs the number of patients in $D$ plus noise from the standard normal distribution is a mechanism.

\textbf{Definition 2:} the \emph{distance of two datasets}, $d(D,D')$, denotes the minimum number of sample changes that are required to change $D$ into $D'$.

For example, if $D$ and $D'$ differ on at most one individual, there is $d(D,D')=1$. We also call such a pair of datasets neighbors. 

The original definition of differential privacy defines as neighbors datasets which `differ on at most one individual'. 
This phrasing has given rise to two different understandings. 
Some interpret this as the replacement of a sample, while others also consider addition and deletion. 
Although the second interpretation is stronger than the first one, 
most of the mechanisms discussed in this paper work for both definitions if slightly modified. 
Different definitions of distance usually lead to different values of sensitivity, while both are bounded. 
In order to make a mechanism designed for one definition of distance work for another definition of distance, we only need to make slight changes to the scale of noise. Therefore we won't distinguish them.

\textbf{Definition 3:} a mechanism $\tilde{f}$ satisfies \emph{($\epsilon,\delta$)-differential privacy} \cite{DBLP:dblp_reference/crypt/Dwork11, DBLP:dblp_conf/eurocrypt/DworkKMMN06} for two non-negative numbers $\epsilon$ and $\delta$ iff for all neighbors $d(D,D')=1$, and all subset $S$ of $\tilde{f}$'s range, as long as the following probabilities are well-defined, there holds
\[
P(\tilde{f}(D)\in S)\leq \delta+e^\epsilon P(\tilde{f}(D')\in S)
\]
Intuitively speaking, the number $\delta$ represents the probability that a mechanism's output varies by more than a factor of $e ^ \epsilon$ when applied to a dataset and any one of its neighbors. A lower value of $\delta$ signifies greater confidence and a smaller value of $\epsilon$ tightens the standard for privacy protection. The smaller $\epsilon$ and $\delta$ are, the closer $P(\tilde{f}(D)\in S)$ and $P(\tilde{f}(D')\in S)$ are, and the stronger protection is. 

There is also a commonly used heuristic to choose $\delta$\cite{DBLP:dblp_conf/kdd/GantaKS08}: when there are $n$ samples in the dataset, $\delta \in o(1/n)$. This is because a mechanism can satisfy $(0,\delta)$-differential privacy but breach privacy with high probability when $\delta$ is large. For each sample in the dataset, the mechanism releases it with probability $\delta$, and the release of different samples are independently. It's easy to prove the mechanism is differentially private. However by expectation, the mechanism release $n\delta$ samples from the dataset. To prevent such leakage, $\delta$ must be smaller than $1/n$.

Typically, ($\epsilon,0$)-differential privacy is simplified to $\epsilon$-differential privacy. With  ($\epsilon,\delta$)-differential privacy, when $\delta > 0$, there is still a small chance that some information is leaked. When $\delta = 0$, the guarantee is not probabilistic. \cite{DBLP:dblp_journals/corr/abs-1107-2183} shows that in terms of mtutal information, $\epsilon$-differential privacy is much stronger than ($\epsilon,\delta$)-differential privacy.

In differential privacy, the number $\epsilon$ is also called the \emph{privacy budget}.

\subsection{Query}
Usually, we want the output of a mechanism to be both differentially private and useful. By `useful', we mean the output accurately answers some queries on the dataset. Definition 4 defines query below and in the following subsections, some mechanisms that guarantee differential privacy will be introduced.

\textbf{Definition 4:} a \emph{query} $f$ is a function that takes a dataset as input. The answer to the query $f$ is denoted $f(D)$.

For example, if $D$ is a medical dataset, then `how many patients were successfully cured?' is a query since it takes $D$ as input and outputs a number. The output of a query is not necessarily a number. However, some mechanisms, notably the Laplacian mechanism, assume that answers to queries are numerical, or vectors $f(D) \in \mathbbm{R}^p$ but not categorical. A more sophisticated query can be `a logistic regression model trained from the dataset', which outputs a classification model.

\subsection{The Laplacian Mechanism}
The Laplacian mechanism\cite{DBLP:dblp_conf/tcc/DworkMNS06} is a popular $\epsilon$-differentially private mechanism for queries $f$ with answers $f(D) \in \mathbbm{R}^p$, in which sensitivity (Definition 5) plays an important role.

\textbf{Definition 5:} given a query $f$ and a norm function $\|.\|$ over the range of $f$, the \emph{sensitivity} $s(f,\|.\|)$ is defined as
\[
s(f,\|.\|)=\max_{d(D,D')=1} \|f(D)-f(D')\|
\]

Usually, the norm function $\|.\|$ is either $L_1$ or $L_2$ norm.

\emph{The Laplacian mechanism\cite{DBLP:dblp_conf/tcc/DworkMNS06}:} given a query $f$ and a norm function over the range of $f$, the random function $\tilde{f}(D)=f(D)+\eta$ satisfies $\epsilon$-differential privacy. Here $\eta$ is a random variable whose probability density function is $p(\eta)\propto e^{-\epsilon\|\eta\|/ s(f,\|.\|)}$.

There is a variation of the Laplacian mechanism, which replaces Laplacian noise with Gaussian noise. On one side, this replacement greatly reduces the probability of very large noise; on the other side, it only preserves ($\epsilon,\delta$)-differential privacy for some $\delta>0$, which is weaker than $\epsilon$-differential privacy.

\emph{Variation of the Laplacian mechanism:} given a query $f$ and a distance function over the range of $f$, the random function $\tilde{f}(D)=f(D)+\eta$ satisfies ($\epsilon,\delta$)-differential privacy. Here $\eta$ is a random variable from distribution $N(0,\frac{2}{\epsilon^2}(s(f,\|.\|))^2\log\frac{2}{\delta})$ \cite{DBLP:dblp_conf/pods/BlumDMN05}.

\subsection{The Exponential Mechanism}
The exponential mechanism\cite{DBLP:dblp_conf/focs/McSherryT07} is an $\epsilon$-differentially private method to select one element from a set. Suppose the set to select from is $A$, and there exists a score function $H$ whose input is a dataset $D$ and a potential answer $a\in A$, and whose output is a real number. Given a dataset $D$, the exponential mechanism selects the element $a\in A$ that has a large score $H(D,a)$.

\textbf{Definition 6:} the \emph{sensitivity} of score function $H$ is defined as
\[
s(H,\|.\|)=\max_{d(D,D')=1,a \in A} \|H(D,a)-H(D',a)\|
\]

\emph{The exponential mechanism:} given a dataset $D$ and a set of possible answers $A$, if a random mechanism selects an answer based on the following probability, then the mechanism is $\epsilon$-differentially private:
\[
P(a \in A \mbox{ is selected})\propto e^{\epsilon H(D,a)/2s(H,\|.\|)}
\]

The Laplacian mechanism is related to the exponential mechanism. If $f(D)$ is a vector in $\mathbbm{R}^p$, and $\forall a\in \mathbbm{R}^p \; H(D,a)=\|a-f(D)\|$, then the output has exactly the same distribution as $\tilde{f}(D)$ in the Laplacian mechanism with half privacy budget.

\subsection{The Smooth Sensitivity Framework and the Sample and Aggregate Framework}
\label{Sec-smooth}
Smooth sensitivity \cite{DBLP:dblp_conf/stoc/NissimRS07}  is a framework which allows one to publish an ($\epsilon,\delta$)-differentially private numerical answer to a query. The noise it adds is determined not only by the query but also by the database itself. By avoiding using the worst-case sensitivity, this framework can enjoy much smaller noise, though the definition of privacy is weaker in this framework compared to the Laplacian mechanism. Two concepts, local and smooth sensitivities, are introduced.

\textbf{Definition 7:} given a query function $f$, a norm function $\|.\|$ and a dataset $D$, the \emph{local sensitivity} of $f$ is defined as:
\[
LS(f,\|.\|,D)=\max_{D':d(D,D')=1} \|f(D)-f(D')\|
\]

One intuitive mechanism would be to add noise to the answer $f(D)$ proportional to the local sensitivity given $(f, D)$. 
However such a mechanism may leak information.
For example, assume $d(D,D')=1$, 
If $D$ has very small local sensitivity and $D'$ has large local sensitivity w.r.t. some query $f$,
the answer given by the mechanism on dataset $D$ is very close to $f(D)$.
However, the answer given $D'$ might be far away from $f(D)$. 
In that case, attackers can infer whether the dataset is $D$ or $D'$ according to the distance between $f(D)$ and the output. To overcome this problem, the smooth sensitivity framework smooths the scale of noise across neighboring datasets.

\textbf{Definition 8:} Given a query function $f$, a norm function $\|.\|$, a dataset $D$ and a number $\beta$, the \emph{$\beta$ smooth sensitivity} of $f$ is defined as
\[
s(f,\|.\|,D,\beta)=\max_{\mbox{Any dataset } D^*} (e^{-\beta d(D,D^*)}LS(f,\|.\|,D^*))
\]

\emph{Smooth sensitivity framework:} Given a query function $f$, the dimension $d$ of the sample space, and Gaussian noise $Z \sim N(0,1)$, the output $\tilde{f}(D)=f(D)+\frac{s(f,\|.\|,D^*,\beta)}{\alpha} Z$ is ($\epsilon,\delta$)-differentially private, provided that $\alpha=\epsilon/\sqrt{\ln(1/\delta)}$ and $\beta=\Omega(\epsilon/\sqrt{d\ln (1/\delta)})$.

The \emph{sample and aggregate framework} \cite{DBLP:dblp_conf/stoc/NissimRS07} is a mechanism 
to respond to queries 
whose answers can be approximated well 
with a small number of samples,
while ensuring ($\epsilon,\delta$)-differential privacy.
The algorithm consists of a sampling step and an aggregating step. 
In the sampling step, the framework partitions the private data set $D$ 
into many subsets $\{D_1,...,D_k\}$, and the answer $f(D_i)$ is estimated on each subset. 
Given the assumption that $f$ can be measured well with small subsets, 
$f(D_1),...,f(D_k)$ are fairly accurate. 
However, we haven't yet placed a privacy constraint on the estimation, 
thus the estimates cannot be released.

In the aggregating step, the framework first defines a quantity $r(i)$ to denote the distance between $f(D_i)$ and $f(D_i)$'s $t$-th nearest neighbor among $f(D_1),...,f(D_k)$ while $t\approx k/2$. Then the framework defines a function $g(f(D_1),...,f(D_k))$ which outputs the $f(D_i)$ with the smallest $r(i)$. Then the smooth sensitivity framework is applied to the function $g(f(D_1),...,f(D_k))$ to ensure differential privacy.

As changing one sample affects only one estimate, the function $g(f(D_1),...,f(D_k))$ has small local sensitivity. Therefore the noise required is small. Furthermore, as most of the estimates $f(D_1),...,f(D_k)$ are close to the true answer, $g(f(D_1),...,f(D_k))$ is accurate. Together, these two properties ensure that the output is accurate.

An efficient aggregation function is provided in the paper \cite{DBLP:dblp_conf/stoc/NissimRS07}. Given $m$ answers $\{f_1,f_2,...,f_m\}$ and a constant $t_0$, the function first computes a quantity $r_i$ for each $f_i$. The quantity $r_i$ is radius of the smallest ball that is centred at $f_i$ and covers at least $t_0$ answers in $\{f_1,f_2,...,f_m\}$. Then the function outputs the answer $f_i$ which has the smallest $r_i$.

\subsection{Combination of Differentially Private Mechanisms}
Sometimes we need to combine several differentially private mechanisms in data processing, thus we need to know how the combination affects the privacy protection. In this subsection, $\tilde{f}_i$ represents differentially private algorithms, $D$ is the dataset, and $\{D_i\}$ is a partition of $D$. Notation $g()$ represents any function. \cite{DBLP:dblp_conf/sigmod/McSherry09} provides the following two theorems.

\emph{Sequential Theorem\cite{DBLP:dblp_conf/tcc/DworkMNS06, DBLP:dblp_conf/eurocrypt/DworkKMMN06, DBLP:dblp_conf/stoc/DworkL09 }:} if $\tilde{f}_i$ is $(\epsilon_i,\delta_i)$-differentially private, then $\tilde{f}(D)=g(\tilde{f}_1(D),\tilde{f}_2(D,\tilde{f}_1(D)),...,\tilde{f}_n(D,\tilde{f}_1(D),\tilde{f}_2(D),...,\tilde{f}_{n-1}(D)))$ is $(\sum_{i=1}^n \epsilon_i,\sum_{i=1}^n \delta_i)$-differentially private.

Intuitively, it means that we can split $\epsilon$ among a sequence of differentially private mechanisms and allow a mechanism in the sequence to use both the dataset and the outputs of previous mechanisms, while the final output is still differentially private. Some more sophisticated forms of this theorem can be find in \cite{DBLP:dblp_conf/focs/DworkRV10, composition2}. 

\emph{Parallel Theorem:} if each $\tilde{f}_i$ is $\epsilon$-differentially private, given a partition $\{D_i\}$ of the dataset $D$, then $\tilde{f}(D=\cup D_i)=g(\tilde{f}_1(D_1),\tilde{f}_2(D_2),...,\tilde{f}_n(D_n))$ is $\epsilon$-differentially private.

If we apply $\epsilon$-differentially private mechanisms to each partition of the dataset, the combined output is still $\epsilon$-differentially private. The partitioning here can be either independent of private data or based on output of some other differentially private mechanism.

Both the sequential method and the parallel method have multiple outputs. A natural question is whether it is always beneficial to the utility of such privacy-preserving mechanisms to average those outputs. The answer is no. For the sequential method, the privacy budget has to be split among several steps; for the parallel method, each partition has less samples than $D$. In both cases, the ratio between the amount of noise applied and the accurate answer is larger than the corresponding ratio for the original mechanism. Therefore, simply averaging them doesn't necessarily lead to better performance.

\section{Machine Learning}

Machine learning algorithms extract information about the distribution of data. Informally, a learning algorithm takes as input a set of \emph{samples} called a \emph{training set} and outputs a model that captures some knowledge about the underlying distribution. Samples are also called \emph{examples}. Typically an \emph{individual} as discussed in the context of differential privacy will correspond to a single \emph{sample} in the machine learning context. The set of all possible samples is called a \emph{sample space}, and all samples in the sample space have the same set of \emph{variables}. These variables can be either categorical or numerical. In the following sections, if there are variables whose values we would like to predict, that are known in training, but unknown for future examples, that variable is denoted $Y$. All the other variables are denoted $\boldsymbol{X}$. When we want to predict labels for new examples, this task is called \emph{supervised learning}. The task in which there are no labels and we want to identify structure in the dataset is called \emph{unsupervised learning}.

The information extracted is represented by machine learning models, and different models are used for different tasks. \emph{Regression} models predict a numerical variable $Y$ given a set of variables $\boldsymbol{X}$. \emph{Classification} models predict a categorical variable $Y$ given a set of variables $\boldsymbol{X}$. \emph{Clustering} models group unlabelled samples into several groups based on similarity. \emph{Dimension reduction} models find a projection from the original sample space to a low-dimensional space, which preserves the most useful information for further machine learning. \emph{Feature selection} techniques select the variables that are most informative for further research. According to whether the learning task is supervised or unsupervised, the training set is denoted either $\{(\boldsymbol{x}_i)\}_{i=1}^n$ or $\{(\boldsymbol{x}_i,y_i)\}_{i=1}^n$, while $n$ is the number of training samples.

Given a family of possible models and a dataset, a machine learning algorithm selects one model that fits the data best. The process of selection is called \emph{training}. 

In the following section, we assume that there is only one variable to predict in regression or classification tasks. For binary classifications, $Y\in \{-1,1\}$. 

In all the following sections, we use the same notation for machine learning tasks. 
Usual capital letters $X$ and $Y$ mean random variables 
while bold capital letters $\boldsymbol{X}$ and $\boldsymbol{Y}$ mean data matrices. 
The $j$-th component of $X$ is denoted $X_j$. 
There are $n$ samples in the dataset 
and the $i$-th sample is denoted $\boldsymbol{x}_i$ for unsupervised learning tasks or $(\boldsymbol{x}_i,y_i)$ for supervised ones. 
The $j$-th component of $\boldsymbol{x}_i$ is denoted $\boldsymbol{x}_{ij}$. 
All constants are denoted by capital letter $C$.

\subsection{Performance Measurement}
Many papers have analyzed the performance of their mechanisms 
and proven that the private models they output are very close to the true models. 
However, the analyses in these papers differ in how they define the true model, 
how they define the distance between two models, 
and given such a distance metric, how they define closeness. 
These differences can impede our efforts to compare different mechanisms.

To assess the performance of a differentially private algorithm, it is necessary to have some notion of a `true model' against which comparisons can be made.
Some papers \cite{DBLP:dblp_conf/stoc/Smith11,DBLP:dblp_journals/corr/abs-0911-5708,DBLP:dblp_journals/jmlr/ChaudhuriMS11,DBLP:dblp_journals/jmlr/JainKT12,DBLP:dblp_conf/icml/0002T13,DBLP:dblp_journals/corr/abs-1211-0975,DBLP:dblp_conf/soda/KapralovT13,DBLP:dblp_conf/nips/ChaudhuriSS12,DBLP:dblp_conf/pkdd/Vinterbo12,DBLP:dblp_conf/colt/ThakurtaS13} 
consider the `true model' to be the output of a noiseless algorithm on training data. 
However, others \cite{DBLP:dblp_conf/stoc/DworkL09,DBLP:dblp_conf/nips/Lei11, DBLP:dblp_conf/icdm/MirW09}  consider the `true model'  to mean the optimal model if the true distribution were known. 

They also differ on how to define the distance between two models. Some papers \cite{DBLP:dblp_journals/jmlr/ChaudhuriMS11,DBLP:dblp_journals/jmlr/JainKT12,DBLP:dblp_conf/pkdd/Vinterbo12} use the difference of values of the target function. Thus the distance between the private model and the true model is the difference between the values taken by the target functions corresponding to each of the two models. 
Some other papers \cite{DBLP:dblp_conf/stoc/DworkL09,DBLP:dblp_conf/stoc/Smith11, DBLP:dblp_conf/nips/Lei11,DBLP:dblp_conf/icdm/MirW09,DBLP:dblp_journals/corr/abs-1211-0975,DBLP:dblp_conf/soda/KapralovT13,DBLP:dblp_conf/nips/ChaudhuriSS12,DBLP:dblp_conf/colt/ThakurtaS13} 
use the distance of the parameters in private and non-private models 
when the models are parametric and have the same parameters. 
Still others \cite{DBLP:dblp_journals/corr/abs-0911-5708,DBLP:dblp_conf/icml/0002T13} use the distance between the predictions made by private and non-private models at certain points in the sample space.

Finally, they differ on the definition of `closeness'. 
Given a measure of distance between two models, 
some papers \cite{DBLP:dblp_conf/stoc/DworkL09,DBLP:dblp_conf/icdm/MirW09} 
prove that as the number of training examples grows large, the output converges to the true model. 
However they do not provide a guaranteed rate of convergence.  
 Other papers \cite{DBLP:dblp_journals/corr/abs-0911-5708,DBLP:dblp_conf/stoc/Smith11, DBLP:dblp_journals/jmlr/ChaudhuriMS11,DBLP:dblp_conf/icml/0002T13,DBLP:dblp_journals/corr/abs-1211-0975,DBLP:dblp_conf/soda/KapralovT13,DBLP:dblp_conf/nips/ChaudhuriSS12, 
DBLP:dblp_conf/pkdd/Vinterbo12,DBLP:dblp_conf/colt/ThakurtaS13}
give bounds on how fast the output models converge to true models. 
For those which prove bounds on the speed of convergence,
 the convergence is usually measured by ($\alpha,\beta$)-usefulness \cite{DBLP:dblp_conf/stoc/BlumLR08}. 
If the mechanism output $\tilde{f}(D)$ is an ($\alpha,\beta$)-useful answer to $f$ on dataset $D$, 
then with probability $1-\beta$, 
the difference between $\tilde{f}(D)$ and $f(D)$ 
is less than $\alpha$. 
Such mechanisms usually provide a relationship between data size, model settings, $\alpha$ and $\beta$. 
A few papers \cite{DBLP:dblp_conf/nips/Lei11} provide worst case guarantees on the distance, which is equivalent to ($\alpha,0$)-usefulness. 
Yet another paper \cite{DBLP:dblp_journals/jmlr/JainKT12} uses the expectation of difference.

Below, we will describe the utility analysis of various algorithms, but we cannot always compare two mechanisms that differ on some of these aspects. Furthermore, even if we can compare the utility of two mechanisms, one might outperform in some situations while the second outperforms in others. Suppose one is $(\epsilon,\delta)$-differentially private and the other is $\epsilon$-differentially private. If we can tolerate a very small probability that information is leaked, then the first may be better; if we are opposed to taking any risk, the second may be better. Therefore, choice of mechanism can depend on specific applications.

\subsection{General Ideas of Differentially Private Machine Learning Algorithms}
Many differentially private machine learning algorithms can be grouped according to the basic approaches they use to compute a privacy-preserving model. This applies both for supervised and unsupervised learning.

Some approaches first learn a model on clean data, and then use either the exponential mechanism or the Laplacian mechanism to generate a noisy model. 
For example, \cite{DBLP:dblp_conf/webi/VaidyaSBH13,DBLP:dblp_conf/icdm/MirW09, DBLP:dblp_conf/imc/SalaZWZZ11,DBLP:dblp_journals/ml/JiE13,DBLP:dblp_journals/tdp/JiangJWMCO13} 
use the Laplacian mechanism, 
while \cite{DBLP:dblp_conf/nips/ChaudhuriSS12,DBLP:dblp_conf/pkdd/Vinterbo12} use the exponential mechanism. 
For some other approaches that have many iterations or multiple steps, 
the Laplacian mechanism and the exponential mechanism are applied to output parameters of each iteration/step. 
Such approaches include \cite{DBLP:dblp_journals/corr/abs-1211-0975,DBLP:dblp_conf/soda/KapralovT13,DBLP:dblp_conf/icdm/JagannathanPW09,DBLP:dblp_conf/kdd/FriedmanS10,DBLP:dblp_journals/jmlr/JainKT12,DBLP:dblp_conf/kdd/MohammedCFY11,DBLP:dblp_conf/sdmw/XiaoXY10}.

Some mechanisms add noise to the target function and use the minimum/maximum of the noisy function as the output model. These technique is called objective perturbation. Some examples include \cite{DBLP:dblp_journals/pvldb/ZhangZXYW12,DBLP:dblp_conf/nips/ChaudhuriM08, DBLP:dblp_journals/jmlr/ChaudhuriMS11, DBLP:dblp_journals/corr/abs-0911-5708}.

Some mechanisms use the idea of the sample and aggregate framework. They are specially designed for queries that can be measured with a small number of samples. First, they split the dataset into many small subsets. Next, they combine the results from all subsets to estimate a model, adding noise in this aggregation step. Mechanisms that employ this idea include \cite{DBLP:dblp_conf/stoc/NissimRS07,DBLP:dblp_conf/colt/ThakurtaS13}. The linear regression in \cite{DBLP:dblp_conf/stoc/DworkL09} is partially based on this idea.

Some mechanisms explore other ideas. 
For example, \cite{DBLP:dblp_conf/nips/Lei11} partitions the sample space 
and uses counts in each partition to estimate the density function. 
\cite{DBLP:dblp_conf/icml/0002T13} interprets a model as a function 
and uses another function to approximate it by iteratively minimizing the largest distance.

Most output perturbation and objective perturbation mechanisms require a bounded sample space. 
This is because unbounded sample space usually leads to unbounded sensitivity. 
Mechanisms based on the sample and aggregate framework don't have this limitation. 
However most of them use $(\epsilon,\delta)$-differential privacy. 
In practice, if the sample space is unbounded and we want to use $\epsilon$-differential privacy, 
we can simply truncate the values in pre-processing. 
If the rule to truncate is independent of the private data, then the truncation is privacy-safe.

In the next sections, some differentially private machine learning mechanisms are introduced. 
We will briefly introduce the learning models, describe additional conditions assumed by various authors, 
explain how they design the mechanisms, 
and provide some utility analysis. 
However, the computation of sensitivity, 
the proof of differential privacy and some mechanism details won't be discussed here.

\section{Differentially Private Supervised Learning}
Supervised machine learning describes the setting when labels are known for training data, and the task is to train a model to predict accurate labels given a new example. In this section we will describe differentially private versions of commonly used supervised machine learning algorithms. 

\subsection{Naive Bayes Model}
The naive Bayes model is a classifier which predicts label $Y$ according to features in $X$. Given features $X$ and a model, one can compute the conditional probability $P(Y|X)$ for all labels $Y$ and predict the label with largest conditional probability. The naive Bayes model is based on two assumptions. 
The first assumption is that $X_j$ are conditionally independent 
given $Y$, i.e., $P(X_j|Y,X_1,...,X_{j-1})=P(X_j|Y)$. 
This enables us to compute the coefficient of each feature independently. 
The second assumption is that for all numerical features in $X$, 
$P(X|Y)$ is a normal distribution.

Based on the first assumption and Bayes' theorem, the conditional probability is as follows:
\[
P(Y|X_1,...,X_p)\propto P(Y)\prod_{j=1}^p P(X_j|Y)
\]
To train the model, we need to estimate all the $P(Y)$ and $P(X_j|Y)$. The probabilities $P(Y)$ can be estimated by the frequencies of samples with label $Y$ in the training set. For conditional probabilities $P(X_j|Y)$, the training is based on whether $X_j$ is categorical or numerical. If $X_j$ is a categorical feature, for all values $x$ and $y$, we have $P(X_j=x|Y=y)=P(X_j=x,Y=y)/P(Y=y)=\sum_i I[\boldsymbol{x}_{ij}=x]I[y_i=y]/\sum_i I[y_i=y]$. Thus we need counts $\sum_i I[y_i=y]$ and $\sum_i I[\boldsymbol{x}_{ij}=x]I[y_i=y]$ to compute the conditional probabilities. If $X_j$ is numeric, then based on the second assumption, the normal distribution $p(X_j|Y)$ is decided by $E[X_j|Y]$ and $Var[X_j|Y]$. Thus to compute the model we only need the following information: $\sum_i I[y_i=y]$, all $\sum_i I[\boldsymbol{x}_{ij}=x]I[y_i=y]$ for categorical variables and all $E[X_j|Y]$ and $Var[X_j|Y]$ for numerical variables.

An $\epsilon$-differentially private naive Bayes model mechanism is introduced in \cite{DBLP:dblp_conf/webi/VaidyaSBH13}. This mechanism relies on one additional assumption: all values for all features in the dataset are bounded by some known number. If the bound covers most of the Gaussian distribution, then both the bound assumption and Gaussian assumption hold approximately. Therefore the sensitivity of the information that is needed to compute the model can be calculated. The mechanism then adds noise to this information according to the Laplacian mechanism and computes the model. Although no analysis on utility is provided, it is easy to see that the noise on the parameters is $O(1/n\epsilon)$.

Sometimes the (non-private) naive Bayes model is more accurate if we model the continuous features with histograms instead of a Gaussian distribution. 
However, in this case, many histograms may lead to high sensitivity. 
Thus as long as the Gaussian assumption is not far from the truth, 
there is no need to use histograms. 
A good assumption about a distribution can result in good performance. 
If in extreme cases the assumption is too far from the truth, 
we can represent those features with histograms in preprocessing. 
If the rule in preprocessing is independent of the private dataset, 
the preprocessing is privacy-free.

Another question is whether we can use the logarithms of counts $\sum_i I[\boldsymbol{x}_{ij}=x]I[y_i=y]$ here, as we sometimes do in using the standard naive Bayes model. 
Clearly, we can apply the Laplacian mechanism to logarithms, however this change is useless. 
For example, suppose the true count is $c$, the noisy count is $\tilde{c}$ and we add noise to $\log(c+1)$. 
According to the Laplacian mechanism, 
the sensitivity of $\log(c)$ is $\log 2$. 
Thus the probability $P(\tilde{(c)=1|c=0})$ stays the same, 
while probabilities such as $P(\tilde{(c)=9|c=4})$ increases a lot. 
Such transformation cannot reduce noise when the count is small, 
however it increases noise a lot when the count is large. 
Therefore, it is better to add noise to the counts directly.

\subsection{Linear Regression}
\label{Sec-linear}

Linear regression is a technique for predicting numerical values $Y$ in which the value is modelled as a linear combination $\boldsymbol{w}^T X$ of features ${X}$. Here, the vector $\boldsymbol{w}$ contains the weights corresponding to each feature and constitutes the set of parameters which must be optimized during training. To train the model, $\boldsymbol{w}$ is computed by minimizing square loss $\sum_{i} (y_i-\boldsymbol{w}^T\boldsymbol{x}_i)^2$ over the training set. 

\cite{DBLP:dblp_journals/pvldb/ZhangZXYW12} assumes bounded sample space and proposes a differentially private mechanism for linear regression. As the loss function is analytic, the mechanism expands the function with Taylor expansion, approximates it with a low order approximation, and adds noise to the coefficients of the terms. The mechanism then finds the $\boldsymbol{w}$ that minimizes the approximate loss function. As the sensitivities of the coefficients are easy to compute, the Laplacian mechanism can ensure the differential privacy of the noisy approximation. Since no private information is used after adding noise, the output vector $\boldsymbol{w}$ here is also $\epsilon$-differentially private. Furthermore, the model that is decided by $\boldsymbol{w}$ is also differentially private.

\subsection{Linear SVM}
\label{Sec-SVM}


Linear SVM is a linear classifier in which a vector $\boldsymbol{w}$ captures the model parameters. A linear SVM model outputs a score $\boldsymbol{w}^T X$ for features ${X}$ in a sample, and usually uses $sign(\boldsymbol{w}^T X)$ as the label $Y$. The parameter $\boldsymbol{w}$ is computed by minimizing $C\sum_{i} \max(0,1-y_i(\boldsymbol{w}^T\boldsymbol{x}_i))+\boldsymbol{w}^T\boldsymbol{w}/2$, while $C>0$ is an input parameter which sets the strength of prediction error.

Under the assumption that the sample space is bounded, the linear SVM model satisfies the following two conditions. First, it computes $\boldsymbol{w}$ by minimizing a strongly convex and differentiable loss function $L(\boldsymbol{w})$. Second, a change of one sample results in bounded change in $L'(\boldsymbol{w})$. 
For linear SVM and all other models satisfying the two conditions, 
\cite{DBLP:dblp_conf/nips/ChaudhuriM08, DBLP:dblp_journals/jmlr/ChaudhuriMS11} provide an output perturbation mechanism and an objective perturbation mechanism. 
The output perturbation mechanism first trains the model and then adds noise to it. 
The objective perturbation mechanism introduces noise 
by adding a carefully designed linear perturbation item 
to the original loss function. 
The $w$ computed from the perturbed loss function is $\epsilon$-differentially private.

\cite{DBLP:dblp_journals/jmlr/ChaudhuriMS11} also provides 
a performance analysis of the objective perturbation mechanism. 
To achieve $(\alpha,\beta)$-usefulness and $\epsilon$-differential privacy, 
the mechanism needs $O(\frac{\log(1/\beta)}{\alpha^2}+\frac{1}{\epsilon\alpha}+\frac{\log(1/\beta)}{\alpha\epsilon})$ samples. As defined earlier, $(\alpha, \beta)$ usefulness provides a guarantee that with respect to the true loss function, the performance of the private model will be within a distance $\alpha$ of that achieved by the true model with probability greater than $1-\beta$.

\subsection{Logistic Regression}
Logistic regression is model for binary classification. It makes prediction $P(Y=1|{X}) =1/(1+e^{-\boldsymbol{w}^TX})$ given features ${X}$ in a sample. The parameters $\boldsymbol{w}$ are trained by minimizing negative log-likelihood $\sum_{i} \log(1+\exp(-y_i\boldsymbol{w}^T\boldsymbol{x}_i))$ over the training set. 

Regularized logistic regression differs from standard logistic regression in that the loss function includes a regularization term. Its $\boldsymbol{w}$ is computed by minimizing $\sum_{i} \log(1+\exp(-y_i\boldsymbol{w}^T\boldsymbol{x}_i))+\lambda \boldsymbol{w}^T\boldsymbol{w}$ over the training set $\{(\boldsymbol{x}_i,y_i)\}$ while $\lambda >0$ is a hyperparameter which sets the strength of regularization. 

Assuming that the sample space is bounded, the mechanism in \cite{DBLP:dblp_journals/pvldb/ZhangZXYW12} (see Section \ref{Sec-linear}) can be applied to make both models $\epsilon$-differentially private. Furthermore, the output perturbation and objective perturbation mechanism in \cite{DBLP:dblp_conf/nips/ChaudhuriM08, DBLP:dblp_journals/jmlr/ChaudhuriMS11} (see Section \ref{Sec-SVM}) can ensure $\epsilon$-differential privacy for regularized logistic regression.

\subsection{Kernel SVM}
Kernel SVM is a machine learning model that uses a \emph{kernel function $K(,)$}, which takes two samples as input and outputs a real number. Different kernel functions lead to different SVM models. Kernel SVM can be used both for classification and regression. When used to classify a sample with features ${X}$, kernel SVM predicts the label $Y=sign(\sum_{i} w_i K({X},\boldsymbol{x}_i))$; when used for regression, kernel SVM predicts the quantity $Y=\sum_{i} w_i K({X},\boldsymbol{x}_i)$. In both cases $\{(\boldsymbol{x}_i,y_i)\}$ are training samples and $w_i$ are weights in the model to compute. Note the model includes the kernel function $K(,)$, all training data and a vector of weights $\{w_i\}_{i=1}^n$. Although there exist many algorithms to train kernel SVM, I will only address those relevant to current differentially private versions.

Unlike previous models, kernel SVMs contain all the training data. Therefore the techniques required to make differentially private kernel SVM mechanisms are different from those we have already described. In \cite{DBLP:dblp_journals/jmlr/ChaudhuriMS11, DBLP:dblp_journals/corr/abs-0911-5708}, an idea for private kernel SVM is proposed. It works for all translation-invariant kernels, where there exists some function $g(\boldsymbol{x})$ such that $K(\boldsymbol{x}_1,\boldsymbol{x}_2)=g(\boldsymbol{x}_1-\boldsymbol{x}_2) \forall \boldsymbol{x}_1,\boldsymbol{x}_2$. For example, radial basis function kernel is translation-invariant. The basic idea is to approximate kernel functions in the original sample space with a linear kernel in another space, so as to avoid publishing training data. It first constructs a space independent of private training data and then projects data from the original sample space to that space. According to \cite{DBLP:dblp_conf/nips/RahimiR07}, the kernel function of two samples in the original sample space can be approximated by the inner product of their projections in the new space. Thus the kernel SVM model turns out to be a linear SVM model in the new space and we can use private linear SVM mechanisms mentioned above. Furthermore, the non-private projection can be published, thus future data can be projected to the same space and then use the parameters from private linear SVM to predict. In this way, the mechanism transforms a kernel SVM model training problem into a linear SVM training problem. 
To achieve both $(\alpha,\beta)$-usefulness w.r.t. to predictions on any sample and $\epsilon$-differential privacy, this mechanism needs $n=O(\frac{\log^{1.5}(1/\alpha\beta)}{\epsilon\alpha^3})$ samples.

The previous mechanism can not be applied to kernel functions that are not translation-invariant, such as polynomial kernel or sigmoid kernel. Therefore \cite{DBLP:dblp_conf/icml/0002T13} proposes another private kernel SVM algorithm for all RKHS kernels. An RKHS kernel means that there is some function $\phi(\boldsymbol{x})$ that projects $x$ onto another space such that $K(\boldsymbol{x}_1,\boldsymbol{x}_2)$ equals the inner product of $\phi(\boldsymbol{x}_1)$ and $\phi(\boldsymbol{x}_2)$. This mechanism seems similar to the one previously described (where the projection can be seen as an approximate to $\phi(\boldsymbol{x})$), however here projection doesn't need to be explicit.

The Test Data-independent Learner (TTDP) mechanism in \cite{DBLP:dblp_conf/icml/0002T13} publishes a private kernel SVM model satisfying ($\epsilon,\delta$)-differential privacy as follows. Intuitively, it trains a non-private kernel SVM model $f(\boldsymbol{x})$ from the private data and then approximates it in a differentially private way. The private model $g(\boldsymbol{x})$ is trained iteratively. First, the mechanism sets $g(\boldsymbol{x})=0$. Then it computes a non-private model $f(\boldsymbol{x})$. Next, it constructs a finite set $Z$ from the unit sphere that represents the sample space. Each iteration consists of three steps. In the first step, the mechanism selects a point $\boldsymbol{z}\in Z$ where $g(\boldsymbol{z})$ and $f(\boldsymbol{z})$ disagree the most. This selection is based on the exponential mechanism, and $|g(\boldsymbol{z})-f(\boldsymbol{z})|$ is used as the score function. In the second step, a noisy difference $|g(\boldsymbol{z})-f(\boldsymbol{z})|+\eta$ is computed while $\eta$ is Laplacian noise. In the third step, the mechanism tests whether the noisy difference exceeds some threshold. If not, the mechanism proceeds to the next iteration; if it does exceed the threshold, the mechanism updates $g(\boldsymbol{x})$ to be closer to $f(\boldsymbol{x})$ at point $\boldsymbol{z}$. After many iterations, $g(\boldsymbol{x})$ may approximate $f(\boldsymbol{x})$. To achieve both $(\alpha,\beta)$-usefulness w.r.t. to predictions on any sample in the unit ball and $(\epsilon,\delta)$-differential privacy, this mechanism needs $n=O(\frac{\log^3\frac{1}{\delta}\log^{1.5}\frac{1}{\beta}}{\alpha^{1.5}\epsilon^{0.75}})$ samples.

\subsection{Decision Tree Learning}
Learning a decision tree classifier involves partitioning the sample space and assigning labels to each partition. The training algorithm for a decision tree classifier consists of a tree building process and a pruning process. In the tree building process, first the entire sample space and all samples are put in the root partition. Then the algorithm iteratively selects an existing partition, selects a variable based on the samples in that partition and a score function such as information gain or Gini index, and partitions the sample space (and the samples corresponding to that partition) according to the variable selected. If the selected variable is categorical, usually each value of that variable corresponds to a partition; if the variable is numerical, then some thresholds will be selected and the partitioning is based on those thresholds. The partitioning process ends when the spaces corresponding to all partitions are small enough or the numbers of samples in each partition are too small. After building the tree, the pruning process removes unnecessary partitions from the tree and merges their spaces and samples to their parents.

\cite{DBLP:dblp_conf/icdm/JagannathanPW09} proposes an $\epsilon$-differentially private mechanism. The mechanism constructs $N$ decision trees and uses the ensemble to make classification. When constructing a decision tree $T_i$, it randomly partitions the sample space into partitions $P_{i1},...,P_{im_i}$ without using private data and computes $Count_{ijy}$, noisy counts of samples with each label $y$ in partition $P_{ij}$. When predicting the label of a sample $X$, it looks for all the partitions $P_{1a_1},...,P_{Na_N}$ from all trees $T_1,...,T_N$ that include $X$, sums the counts of samples with each label from all those partitions $S_y=\sum_{i=1}^n Count_{ia_iy}$, and computes the probabilities of label $y'$ by $P(Y=y'|X)=S_{y'}/\sum_y S_y$.

\cite{DBLP:dblp_conf/kdd/FriedmanS10} proposes another $\epsilon$-differentially private decision tree algorithm. The mechanism is based on the assumption that all features are categorical, in order to avoid selecting partition points for any feature. In the partitioning process, the mechanism uses the exponential mechanism to select the variable with largest score (for example, information gain or Gini index) differentially privately. Each time a partition reaches a pre-determined depth, or the number of samples in that partition is about the same scale as random noise, or the sample space corresponding to that partition is too small, the mechanism stops operating on that partition. It then assigns to that partition a noisy count of samples with each label. After the partitioning process has completed altogether, these noisy counts are used to decide whether to remove those nodes without having to consider privacy.

\subsection{Online Convex Programming}
Many machine learning techniques, such as logistic regression and SVM, specify optimization problems which must then be solved to find the optimal parameters. Online algorithms, such as gradient descent, which consider examples one at a time, are widely used for this purpose. To that a machine learning algorithm is differentially private, it is therefore important to demonstrate that the optimization algorithm doesn't leak information.

Online convex programming (OCP) solves convex programming problems in an online manner. The input to an OCP algorithm is a sequence of functions $(f_1,...,f_T)$ and the output is a sequence of points $(w_1,...,w_T)$ from a convex set $C$. The algorithm is iterative and starts from a random point $w_0$. In the $t$-th iteration, the algorithm receives the function $f_t$ and outputs a point $w_{t+1}\in C$ according to $(f_1,...,f_t)$ and $(w_1,...,w_t)$. The target of the OCP algorithm is to minimize regret, which is defined as
\[
R=\sum_{t=1}^T f_t(w_t)-\min_{w\in C}\sum_{t=1}^T f_t(w).
\]
Note that the input here is a sequence of functions instead of samples. 

There are many methods to find $w_{t+1}\in C$ according to $(f_1,...,f_t)$ and $(w_1,...,w_t)$. 
\cite{DBLP:dblp_journals/jmlr/JainKT12} provides ($\epsilon,\delta$)-differentially private versions for two of them: the Implicit Gradient Descent (IGD) and the Generalized Infinitesimal Gradient Ascent (GIGA) given all the functions are $L$-Lipschitz continuous for some constant $L$ and $\eta$-strongly convex. 
IGD first computes
\[\hat{w}_{t+1}=\arg\min_{w\in C} \left(\frac{1}{2}\|w-w_t\|^2+\frac{1}{\eta t}f_t(w)\right)\]
and projects $\hat{w}_{t+1}$ onto $C$ to get the output $w_{t+1}$. 
GIGA first computes 
\[\hat{w}_{t+1}=w_t-\frac{1}{\eta t}\nabla f_t(w_t) \] 
and then does the projection. 
Both algorithms ensure bounded sensitivity of $w_{t+1}$ given $w_t$. 
The private mechanism in \cite{DBLP:dblp_journals/jmlr/JainKT12} adds Gaussian noise to every $\hat{w}_t$ before it is projected to $w_t$ to preserve privacy, 
and then use the noisy $w_t$ for the future computation. 
Given $T$ functions(samples), the expected regret by this ($\epsilon,\delta$)-differentially private mechanism is $O\left(\sqrt{\frac{T}{\epsilon}} \ln^2 \frac{T}{\delta}\right)$.

\section{Differentially Private Unsupervised Learning}
Unsupervised learning describes the setting when there are no labels associated with each training example. In the absence of labels, unsupervised machine learning algorithms find structure in the dataset. In clustering, for example, seeks to find distinct groups to which each datapoint belongs. It can be useful in many contexts such as medical diagnosis, to know of an individuals member ship in a group which shares certain specific characteristics.  However, releasing the high-level information about a group may inadvertently leak information about the individuals in the dataset.  Therefore it is important to develop differentially private unsupervised machine learning algorithms.

\subsection{K-means clustering}
K-means is a commonly used model in clustering. To train the model, the algorithm starts with $k$ randomly selected points which represent the $k$ groups, then iteratively clusters samples to the nearest point and updates the points by the mean of the samples that are clustered to the points.

\cite{DBLP:dblp_conf/stoc/NissimRS07} proposes an ($\epsilon,\delta$)-differentially private k-means clustering algorithm using the sample and aggregate framework. The mechanism is based on the assumption that the data are well-separated. `Well separated' means that the clusters can be estimated easily with a small number of samples. This is a prerequisite of the sample and aggregate framework. The mechanism randomly splits the training set into many subsets, runs the non-private k-means algorithm on each subset to get many outputs, and then uses the smooth sensitivity framework to publish the output from a dense region differentially privately. This step preserves privacy while the underlying k-means algorithm is unchanged.

Any modifications on the k-means clustering algorithm (such as k-means++) can be used in the sample step, with the sole restriction that such modifications leave intact the property that the algorithm can be estimated with a small number of samples. Additionally, if the sample space is bounded and the number of samples surpasses a threshold, there is a bound on the noise added. However this bound is not directly related to the number of samples in the dataset.

\section{Differentially Private Dimensionality Reduction}
In machine learning contexts, when data is high dimensional, it is often desirable learn a low-dimensional representation. Lower dimensional datasets yield models with less degrees of freedom and tend to be less prone to overfitting. From a differential privacy perspective, lower dimensional representations are desirable because they tend to have lower sensitivity. 

Feature selection is one technique for dimensionality reduction, in which a subset of features is kept from an original feature space. Principal component analysis (PCA), on the other hand is a matrix factorization technique in which a linear projection of the original dataset into a low dimensional space is learned such that the new representation explains as much of the variance in the original dataset as possible.

\subsection{Feature Selection}
\cite{DBLP:dblp_conf/pkdd/Vinterbo12} proposes an $\epsilon$-differentially private feature selection, PrivateKD, for classification. PrivateKD is based on the assumption that all features are categorical and each feature has finite possible values. For any set of features $S$, it defines a function $F(S)$ which tells how many pairs of samples from different classes can features in $S$ distinguish. The set of selected features $S'$ is initialized to $\emptyset$. Then a greedy algorithm adds new features one by one to $S'$. When selecting a feature to add, the algorithm uses the exponential mechanism to select the feature that can lead to the largest increase of $F(S')$. The paper provides a utility guarantee for the special case where the cardinality of sample space $m$ and the number of features $d$ have the relation $m=d-1$. In that case, except probability $O(1/poly(m))$ ($poly(m)$ means a polynomial expression of $m$), $F(S')\geq (1-1/e)F(S_{optimal})-O(\log m/\epsilon)$.

\cite{DBLP:dblp_conf/colt/ThakurtaS13} proposes an ($\epsilon,\delta$)-differentially private algorithm for feature selection when the target function is stable. Unlike the previous paper, this paper doesn't explicitly state the algorithm for feature selection. Instead, it only requires the selection algorithm to be stable. By `stable', we mean that either the value of function as calculated on the input dataset doesn't change when some samples in the set change, or that the function can output the same result on a random subset from the input dataset with high probability. For the first kind of functions, the mechanism uses the smooth sensitivity framework in \cite{DBLP:dblp_conf/stoc/NissimRS07} to select features. If adding or removing any $\frac{\log(1/\delta\beta)}{\epsilon}$ samples from the input dataset doesn't change the selection result, then the algorithm can output the correct selection result with probability $1-\beta$.

For the second kind of functions, the mechanism uses an idea similar to the sample and aggregate framework in \cite{DBLP:dblp_conf/stoc/NissimRS07}: it creates some bootstrap sets from the private dataset, selects features non-privately on each set, and counts the frequencies of feature sets output by the algorithm. Intuitively, if the number of samples is large and the features set is not too large, there is high probability that the correct output is far more frequent than any other one. Thus, the mechanism can release the most frequently selected set of features. If a random subset of the input dataset with $\epsilon/(32\log(1/\delta))$ samples outputs the same selection result with probability at least $3/4$, then the mechanism outputs the correct solution with probability at least $1-\delta$.

\subsection{Principal Component Analysis}
Principal Component Analysis (PCA) is a popular method in dimension reduction. 
It finds $k$ orthogonal directions on which the projections of data have largest variance. 
The original data can then be represented by its projection onto those $k$ directions. 
Usually $k$ is much smaller than the dimension of sample space.
Thus the projection greatly reduces the dimensionality of the data. 
It is well-known that this analysis is closely related to eigen-decomposition: 
if we rank the eigenvectors of matrix $\boldsymbol{A}=Var[{X}]$ 
according to the corresponding eigenvalues $\lambda_1\geq\lambda_2\geq...\geq\lambda_p$, 
then the first $k$ eigenvectors are the $k$ directions.

There are two differentially private mechanisms to select eigenvectors iteratively. 
The iterative methods are based on the spectral decomposition, 
which ensures that if the components corresponding to the first $i-1$ eigenvectors of $\boldsymbol{A}$ are subtracted from $\boldsymbol{A}$, 
then the $i$-th largest eigenvector becomes the largest eigenvector of what remains. 
Therefore the process of selecting the largest $k$ eigenvectors 
can be replaced by repeatedly finding the first eigenvector 
and removing the component corresponding to the selected eigenvector. 
The following two mechanisms both make use of this idea 
but differ on how to select the first eigenvector.

An ($\epsilon,\delta$)-differentially private mechanism is proposed in \cite{DBLP:dblp_journals/corr/abs-1211-0975}. 
The mechanism uses the power method: $\boldsymbol{A}^n \boldsymbol{v}/\|\boldsymbol{A}^n\boldsymbol{v}\|$ converges 
to the first eigenvector of $\boldsymbol{A}$ if $\boldsymbol{v}$ is not orthogonal to the first eigenvector. 
It randomly starts with a unit-length vector $\boldsymbol{v}$, 
then iteratively updates $\boldsymbol{v}$ with $(\boldsymbol{Av}+\boldsymbol{\eta}_i)/\|\boldsymbol{Av}+\boldsymbol{\eta}_i\|$ while $\boldsymbol{\eta}_i$ is Gaussian noise in the $i$-th iteration. 
Since it is exceedingly improbable that a random vector is orthogonal to the first eigenvector, 
the vector $\boldsymbol{v}$ will get close to the first eigenvector. 
However due to the noise, $\boldsymbol{v}$ cannot converge with arbitrary accuracy. 
Thus it outputs $\boldsymbol{v}$ after a fixed number of iterations and proceed to find the next largest eigenvector. 
A utility guarantee is provided on the power method, 
which outputs the first eigenvector. 
However there is no direct guarantee on the $k$ eigenvectors. 
For each eigenvector $\boldsymbol{a}$ output by running the power method on matrix $\boldsymbol{A}$, 
the distance from the first eigenvector ($\|\boldsymbol{Aa}\|/\|\boldsymbol{a}\|-\lambda_1$ while $\lambda_1$ is the first eigenvalue) is $O((\sqrt{\log(1/\delta)}\log n)/\epsilon)$.

\cite{DBLP:dblp_conf/soda/KapralovT13} provides an $\epsilon$-differentially private mechanism for principal component analysis. 
According to the property that the first eigenvector $\boldsymbol{v}$ of $\boldsymbol{A}$ is the unit-length vector that maximizes $\boldsymbol{v}^T\boldsymbol{Av}$, 
the mechanism uses $H(\boldsymbol{X},\boldsymbol{v})=\boldsymbol{v}^T\boldsymbol{Av}$ as the score function in the exponential mechanism to select the first eigenvector from the set $\{\boldsymbol{v}:\boldsymbol{v}^T\boldsymbol{v}=1\}$ differentially privately. 
The selection algorithm is specially designed to be computable in reasonable time. 
This paper also provides two proofs on utility of this mechanism. 
For any $0<\delta<1$ and privacy budget $\epsilon$, 
if the first eigenvalue of matrix $\boldsymbol{A}$, 
$\lambda_1>O(\ln(1/\delta)/(n\epsilon\delta))$, 
then the first eigenvector $\boldsymbol{v}$ has the property $E[\boldsymbol{v}^T\boldsymbol{Av}]\geq (1-\delta)\lambda_1$. 
For any $0<\delta<1$ and privacy budget $\epsilon$, 
if the first eigenvalue $\lambda_1>O(1/(n\epsilon\delta^6))$, 
the $k+1$-th eigenvalue is denoted $\lambda_{k+1}$, 
and the $k$-rank approximation matrix output is denoted $\boldsymbol{A}_k$, 
then the largest eigenvalue of $\boldsymbol{A}-\boldsymbol{A}_k$ is smaller than $\lambda_{k+1}+\delta\lambda_1$ with large probability.

Not all differentially private approaches to PCA rely on iterative algorithms. 
\cite{DBLP:dblp_conf/nips/ChaudhuriSS12} proposes an $\epsilon$-differentially private mechanism, PPCA, 
to compute $k$ largest eigenvectors at the same time. 
The mechanism uses the property that the first $k$ eigenvectors of $\boldsymbol{A}$ 
are the columns of the $p\times k$ matrix $\boldsymbol{V}=\arg\max_{\boldsymbol{V}:\boldsymbol{V}^T\boldsymbol{V}=\boldsymbol{I}_k} tr(\boldsymbol{V}^T\boldsymbol{AV})$. 
Therefore, it uses $H(\boldsymbol{X},\boldsymbol{V})=tr(\boldsymbol{V}^T\boldsymbol{AV})$ as the score function 
and selects $\boldsymbol{V}$ from the set of matrices $\{\boldsymbol{V}:\boldsymbol{V}^T\boldsymbol{V}=\boldsymbol{I}_k\}$ 
according to the exponential mechanism. 
A Gibbs sampler method is used here to select the matrix $\boldsymbol{V}$. 
This paper provides a guarantee on the first eigenvector. 
For any $0<\rho,\eta<1$, if the sample size $n=O\left(\frac{1}{\epsilon(1-\rho)}\left(\log\frac{1}{\eta}+\log\frac{1}{1-\rho^2}\right)\right)$, 
then the inner product of output first eigenvector and true eigenvector 
is larger than $\rho$ with probability $1-\eta$.

\section{Statistical Estimators}
Statistical estimators calculate approximations of quantities of interest based upon the evidence in a given dataset. Simple examples include the population mean and variance. While estimators are clearly useful, they may potentially leak information about the individuals contained in the dataset, especially when the dataset is small or features are rare. Therefore, to protect privacy, it is necessary to develop differentially private estimators.

\subsection{Robust Statistics Estimator}
\label{Section-rse}
\cite{DBLP:dblp_conf/stoc/DworkL09}  proposes an $(\epsilon,\delta)$-differentially private mechanism for robust statistical estimators. Roughly speaking, a statistical estimator produces an estimate of a vector (such as the mean and variance of a Gaussian distribution) based on the input dataset. The estimator $T$ can be seen as a function that maps a dataset $D$ to the output vector $T(D)$. Most statistical estimators converge when the number of samples tends to infinity and the samples are i.i.d. drawn from some distribution $P$. When the estimator does converge, the limit $\lim_{|D|\rightarrow +\infty} T(D)$ is denoted $T(P)$. The definition of robust estimator is based on the stability of estimates. An estimator is robust if for any element $\boldsymbol{x}$ in the sample space, the following limit exists $\lim_{t\rightarrow 0} (T((1-t)P+t\delta_{\boldsymbol{x}})-T(P))/t$. The distribution $(1-t)P+t\delta_{\boldsymbol{x}}$ means that with probability $1-t$ the sample is from $P$ and with probability $t$ the sample is $\boldsymbol{x}$.

The output of a robust estimator doesn't change much if a small number of samples change. Based on the property, \cite{DBLP:dblp_conf/stoc/DworkL09} comes up with a Propose-Test-Release framework. The framework is based on the assumption that the statistics are in $\mathbbm{R}^p$. It divides $\mathbbm{R}^p$ into small cubes, then computes the statistics $T(D)$ from a dataset $D$ and the number of sample changes needed to make $T(D)$ fall into another cube. If the number is large, the statistics are stable and thus the mechanism can add Laplacian noise to the statistics to make it private; if the number is small, then the mechanism outputs $\perp$, which means it fails. When the number of samples tends to infinity, and the samples are i.i.d. drawn, the framework output is asymptotically equivalent to a non-private robust estimator.

Based on this framework, \cite{DBLP:dblp_conf/stoc/DworkL09} proposes three mechanisms for interquartile range estimation, trimmed mean and median, and linear regression, respectively. When applying the framework to linear regression, the framework uses a robust estimator to learn a model from the training set $\{(\boldsymbol{x}_i,y_i)\}$
\[
\hat{\boldsymbol{w}} =\arg\min_{\boldsymbol{w}} \sum_{i} \frac{|y_i-\boldsymbol{w}^T \boldsymbol{x}_i|}{\|\boldsymbol{x}_i\|}
\] 
instead of minimizing the mean square error. Given $n$ samples, 
the linear regression estimator can successfully output a model with probability $1-O(n^{-c\ln n})$ for some constant $c$. Additionally, its output converges to the true linear regression parameter when $n$ tends to infinite.

\cite{DBLP:dblp_conf/icml/ChaudhuriH12} explores robust estimators in another way. 
They prove that if effect of one sample is bounded by $O(1/n)$, 
and the range of $T(P)$ is bounded, 
then the smooth sensitivity framework provides bounded error. 

However, if $T(P)$ is not bounded, 
and if for any value $\tau$ in an infinite range, 
there exists some $P$ such that $T(P)=\tau$, 
then the error of any $\epsilon$-differentially private mechanism cannot be upper bounded. 

Here, a bound on the effect of one sample means 
that there exists a uniform upper bound $M(P)$ for distribution $P$, 
such that for all distributions $P'$ satisfying $|P-P'|\leq O(\sqrt{1/n})$, 
all $x$ in the sample space, 
$T((1-1/n)P+\delta_{\boldsymbol{x}}/n)-T(P))\leq M(P)/n$. 

To achieve both $(\epsilon,\delta)$-differential privacy and $(\alpha, \beta)$-usefulness, 
the smooth sensitivity framework requires $O\left(\frac{\ln(1/\beta)}{\epsilon\alpha}+\frac{\ln^2(1/\delta)\ln^2(\alpha\epsilon/\ln(1/\beta))}{\epsilon^2\ln(1/\beta)}\right)$ samples.
\normalsize

\subsection{Point Estimator}
\cite{DBLP:dblp_journals/corr/abs-0809-4794, DBLP:dblp_conf/stoc/Smith11} give a differentially private mechanism for point estimation. Using the notation in \ref{Section-rse}, the mechanism splits $D$ into $k$ subsets 
with equal size $\{D_1...,D_k\}$ randomly 
and estimates parameters $\{T(D_1),T(D_2),...,T(D_k)\}$ on each subset. 
Then it uses a differentially private mean of all $T(D_i)$ to approximate $T(D)$. 
The mean is computed in two steps. 
First, if the space of parameters is unbounded, 
it computes two quantiles 
and truncates all estimates according to the quantiles and the sample size. 
Second, the mechanism adds Laplacian noise to the mean of truncated values 
and publishes the noisy mean. 
When the space of possible parameters are bounded, 
then the mechanism only executes the second step 
and becomes $\epsilon$-differentially private. 
When the space of possible parameters is not bounded, 
then both steps have to be executed and the mechanism is $(\epsilon,\delta)$-differentially private.

The papers give sufficient conditions under which the mechanism 
is as accurate as non-private estimators asymptotically. 
The conditions are listed below. 
$n$ is the number of samples in $D$ and $\sigma_P$ is a real number. 
For models that don't converge to a fixed point (for example, some EM algorithms) 
or models with varying numbers of features (such as Kernel SVM), 
there is no such guarantee.
\[
\frac{T(D)-T(P)}{\sigma_P/\sqrt{n}}\rightarrow N(0,1) \mbox{ when } n\rightarrow +\infty
\]
\[
E[T(D)]-T(P)=O(1/n)
\]
\[
E\left[\left(\frac{|T(D)-T(P)|}{\sigma_P/\sqrt{n}}\right)^3\right]=O(1)
\]

\subsection{M-estimator}

\cite{DBLP:dblp_conf/nips/Lei11} proposes an $\epsilon$-differentially private mechanism for M-estimator. 
Unlike the robust estimator above, 
the definition of an M-estimator depends on the function from which the estimates come. 
M-estimation relies on a function $m(,)$, 
which takes a sample and a parameter $\theta$ as input 
and outputs a real number. 
An M-estimator estimates the parameter $\theta$ by computing
\[
\hat{\theta}=\arg\min_{\theta} \frac{1}{n}\sum_i m(\boldsymbol{x}_i,\theta)
\]
The mechanism in \cite{DBLP:dblp_conf/nips/Lei11} 
first divides the sample space ($[0,1]^d$) into many small cubes without using private data. 
Then it adds Laplacian noise to the counts of samples in the cubes 
and computes the density function in each cube 
by dividing the noisy count corresponding to that cube by the volume of the cube. 
The noisy density function leads to a noisy target function in the minimization problem above, 
and the noisy target function leads to a noisy minimum $\tilde{\theta}$. 
The noisy minimum can be released as an estimate.

The output parameter will converge to the true parameter based on the distribution of training data with speed $O(n^{-1/2}+(\sqrt{\log n}/n)^{2/(d+2)})$ under some regularity conditions. Here $d$ is the number of features and $n$ is the number of samples.

\section{Learning in Private Data Release}

Many differentially private data release mechanisms have been described, such as \cite{DBLP:dblp_conf/stoc/BlumLR08,DBLP:dblp_conf/pods/BarakCDKMT07, DBLP:dblp_conf/pet/FangC12, DBLP:dblp_conf/icdt/CormodePST12}. 
In this section, we focus on data release mechanisms 
that are either useful for machine learning 
or based on machine learning algorithms.

Many papers use partition based algorithms to release data. 
\cite{DBLP:dblp_conf/sdmw/XiaoXY10} assumes that the density function is smooth, 
\cite{DBLP:dblp_journals/pvldb/ChenMFDX11, DBLP:dblp_conf/dasfaa/ZhangMC13} assume that the data's format permits it to be organized in a tree, 
and  \cite{DBLP:dblp_conf/kdd/MohammedCFY11} assumes 
that partitions can preserve most important information for further data mining. 
All these assumptions motivate partitioning the sample space and publishing counts in each partition. 
With respect to the mechanism design, the mechanisms can be divided into two groups. 
\cite{DBLP:dblp_conf/kdd/MohammedCFY11} first partitions the sample space using the exponential mechanism and then adds noise to the counts using the Laplacian mechanism. 
Some  others (\cite{DBLP:dblp_journals/pvldb/ChenMFDX11, DBLP:dblp_conf/dasfaa/ZhangMC13, DBLP:dblp_conf/sdmw/XiaoXY10}) generate noisy counts with the Laplacian mechanism for each cell and then partition according to the noisy counts.


Some  data release mechanisms don't depend on partitioning. 
Some of them assume the private data is fit well by some family of models.
They select an optimal model from that family privately 
and then generate new data according to the selected model. 
Some others assume some property (like sparsity) of the data, 
and propose mechanisms that can make use of that property.

\cite{DBLP:dblp_conf/icdm/MirW09, DBLP:dblp_conf/imc/SalaZWZZ11} publish a graph generator model based on the assumption that the private data is fit well by some parametrized generative model. 
Though the two mechanisms use different generative models, 
both train the model first, 
add Laplacian noise to the parameters of the model, 
and then use the noisy model to generate a new graph.

\cite{DBLP:dblp_conf/kdd/XiaoCT14} represents the network structure by using a statistical hierarchical random graph model. 
Unlike the two models above, 
the number of parameters in this model is proportional to the number of nodes in the graph. 
Thus we will introduce too much noise if we use the Laplacian mechanism to publish the model. 
As the parameter space is very large 
and no score function exists, 
which is both simple and meaningful, 
it is not easy to use the exponential mechanism directly. 
To overcome this difficulty, 
the authors propose a mechanism based on the Markov Chain Monte Carlo procedure. 
It uses the Metropolis–Hastings algorithm 
to draw a model from the distribution in the exponential mechanism 
given the score function is the likelihood function. 
Though the likelihood function is complicated w.r.t. the whole space of parameters, 
it is simple w.r.t. one parameter if all the others are fixed. 
Thus the Markov Chain Monte Carlo procedure is possible. 
The graph can be reconstructed from the output model after many iterations.

Sometimes corresponding to a large private dataset of interest, 
there exists a similarly structured but smaller publicly available dataset. 
\cite{DBLP:dblp_journals/ml/JiE13} makes use of a public dataset, 
assigning weights to its examples to embed information contained in the private data. 
The mechanism assumes the existence of such a public dataset, 
but does not require that it is drawn from the same distribution as the private data. 
First they use public/private datasets as positive/negative samples in a logistic regression model. 
They train a noisy logistic regression model and assign weights based on the noisy model to all public samples. 
The weighted public dataset can replace the private dataset in future data mining. 
If the weighted set is used in measuring the expectation $E[f(X)]$ for some function $f(X)$ while $X$ is from the distribution of private data, the standard deviation of the estimate is $O(1/\sqrt{n})$. 
If the assumption holds exactly,
then the estimate is asymptotically unbiased.

\cite{DBLP:dblp_journals/tdp/JiangJWMCO13} assumes that samples have binary labels 
and that they are fit by a linear discriminant analysis (LDA) model. 
The idea is similar to the exponential mechanism: 
The exponential mechanism computes the non-private LDA model from the private dataset $D$. 
Then, it uses the distance between the non-private model parameters 
and the LDA parameters trained from another dataset $D'$ as the score function, 
and draws a dataset differentially privately.

The mechanism in this paper, however, first computes a private LDA model by the Laplacian mechanism, 
and then draws a dataset that minimizes the distance. 
Such an output dataset can preserve the classification information from the private data.

\cite{DBLP:dblp_conf/wpes/LiZWY11} assumes that the data matrix is sparse, 
thus the mechanism can make use of results from compressive sensing research.
Informally speaking, 
if we randomly project the high-dimensional data matrix to a low-dimensional space, 
and then attempt to recover the high-dimensional matrix using the low-dimensional embedding 
and the constraint that the recovered matrix is sparse, 
there is high probability that the recovered matrix exactly matches the original one. 
\cite{DBLP:dblp_conf/wpes/LiZWY11} first randomly projects the data matrix, 
then adds noise to the compressed information, 
and reconstructs data from the noisy compressed information. 
As the dimension of compressed information is much smaller than that of the original data matrix, 
the scale of Laplacian noise needed to preserve $\epsilon$-differential privacy can be reduced from $O(\sqrt{n}/\epsilon)$ to $O(\log n/\epsilon)$ given $n$ samples. 

\section{Theoretical Results}
\cite{DBLP:dblp_conf/focs/KasiviswanathanLNRS08} studies the general properties of private classifiers, 
instead of individual learning models. 
They first define a problem to be learnable 
if there is an algorithm that can output a highly accurate model 
with a large probability given enough data. 
The accuracy here is measured as the percentage of samples that are correctly classified. 
They further claim that if a problem is learnable, 
then it can be learned differentially privately. 
The mechanism they constructs takes the number of correctly classified samples as a score function 
and uses the exponential mechanism to draw the best model, 
which it calls the best hypothesis in a class.

\cite{DBLP:dblp_conf/edbt/Mir12} formulates differentially private learning in an information theoretic framework. 
The paper uses a concept in information theory, PAC-Bayesian bound. 
This concept is for parametrized models that have bounded loss functions 
and it uses the Bayesian learning framework. 
PAC-Bayesian bound $PAC(\tilde{\pi},\pi,\epsilon,\lambda)$ is a function of the posterior distribution $\tilde{\pi}$ of model parameters, 
the prior distribution $\pi$ of parameters, 
a positive number $\epsilon$ representing the privacy budget,
and another number $\lambda\in(0,1)$, which is related to the strength of the bound. 
With probability $1-\lambda$ and prior distribution $\pi$, 
$PAC(\tilde{\pi},\pi,\epsilon,\lambda)$ upper bounds the expected loss on the true distribution 
if the model parameter is from the distribution $\tilde{\pi}$. 
This bound varies given different $\epsilon$. 
According to \cite{DBLP:dblp_conf/edbt/Mir12}, 
the output of the exponential mechanism 
follows the posterior distribution that minimizes PAC-Bayesian bound.

Note that the PAC-Bayesian bound upper bounds the loss function. 
It is possible that other mechanisms have loss functions that achieve a better bound. 
Therefore the conclusion in \cite{DBLP:dblp_conf/edbt/Mir12} doesn't necessarily mean that the exponential mechanism is the best.
\input{discussion.tex}

\section*{Acknowledgments}
The authors would like to thank Kamalika Chaudhuri for her comments. 
The authors are in part funded by NLM(R00LM011392).

\bibliography{TKDE}
\bibliographystyle{plain}

\end{document}

%% file: discussion.tex
\section{Discussion}

The papers reviewed by this survey address
the question of how to train a differentially private model 
with as little noise as possible. 
To summarize, there are generally four guiding principles for reducing the scale of noise.
First, adding noise only one time is usually better than adding noise many times. 
This is because if we add noise many times, 
we have to split the privacy budget into many smaller portions 
and let each noise addition procedure use one portion. 
Because the budget allocated to each procedure is small 
and the scale of noise is inversely related to the privacy budget, 
the amount of noise added in each procedure is large. 
Furthermore, when we aggregate the outputs, the noise can grow even larger.
Therefore one-time noise addition is usually better.

For example, when we train a logistic regression model, 
we can add noise to the training process, the target function or the final model. 
Adding noise to the target function is a one-time procedure.
This is also true for the final model. 
However, as the training process is iterative, 
adding noise during training requires adding noise many times. 
According to our experience, 
noise addition in training process leads to significantly worse performance.

Second, lower global sensitivity (compared to the result) leads to smaller noise.
In one strategy to lower global sensitivity, some queries be approximated by combining the results of other queries, each of which have far lower global sensitivities than the original query. 
For example, \cite{DBLP:dblp_conf/webi/VaidyaSBH13} adds noise to the counts 
that generate the naive Bayes model 
instead of conditional probabilities of the model directly. 
The global sensitivity of each conditional probability is 1, which is too high to be useful. The global sensitivity of each underlying count is 1, which is much lower compared to the counts. 
By adding noise to those counts, we encounter lower global sensitivity.

Another approach is to modify the model. 
For example, \cite{DBLP:dblp_journals/corr/abs-0911-5708} transforms kernel SVM to linear SVM, 
and \cite{DBLP:dblp_conf/stoc/DworkL09} uses a robust linear regression model 
to replace the commonly used model. 

Third, use of public data, when available, can reduce the noise in some cases. 
For a private dataset, 
there is often a smaller public dataset drawn from a similar population. 
This public dataset can be from a previous leak or by consent of data owners. 
Because differentially private mechanisms distort private data, 
the smaller public dataset sometimes provides similar or better utility. 
According to \cite{DBLP:dblp_journals/ml/JiE13, parlog}, 
such a public dataset can enhance the performance of differentially private mechanisms. 

Fourth, for some models, 
iterative noise addition may be reasonable.
There are some times when the sensitivity of output model parameters is very large 
but the iterative algorithm has smaller sensitivity. 
This statement may seem counterintuitive, 
as the sum of sensitivities of all iterations 
should be similar to the sensitivity of the model parameters. 
However, in some cases, the sensitivity of each iteration 
is determined by the parameter before that iteration. 
Thus the sum of sensitivities of those iterations in fact relies on the training path. 
Excepting some extreme cases, 
the sum can be much smaller than the sensitivity of the model parameters. 
In this case, it seems necessary to add noise in the iterations. 

For those models, one can consider trying the MCMC-based algorithm as in \cite{DBLP:dblp_conf/kdd/ShenY13}. 
Likelihood function or loss function can be used as score functions 
and the Metropolis Hastings algorithm ensures that the output 
is from the same distribution as that in the exponential mechanism. 
This idea is still not widely used, 
however it seems possible that it improves learning performance.

In addition to these four ideas, some other issues warrant attention.
For example, most differentially private mechanisms use clean and complete data as input, 
which is not always available in practice. 
Furthermore, traditional methods for missing data or pre-processing may not satisfy differential privacy. 
Thus mechanisms that can deal with incomplete data are desired. 
Such mechanisms can either release data, 
or be combined with other differentially private learning mechanisms.

When private data is discussed, 
medical data is typically offered as an example application. 
However, medical datasets are often not relational. 
They may be temporal, and sometimes structural. 
Although we can transform such data, 
the transformation may lose some important information and increase sensitivity. 
Therefore, mechanisms specially designed for such data are required.

Another important question is whether privacy can be free, i.e., 
achieved at no cost to utility
in differentially private learning. 
For privacy to be free, the noise required to preserve privacy 
might need to be smaller than noise from sample randomness. 
In that case, it wouldn't change the magnitude of noise to take privacy into account. 
For example, \cite{DBLP:dblp_conf/stoc/Smith11} proves that $(\epsilon,\delta)$-differential privacy 
is free for learning models satisfying a certain set of conditions. 
The mechanism in \cite{DBLP:dblp_journals/jmlr/ChaudhuriMS11} 
ensures free $\epsilon$-differential privacy for regularized logistic regression models and linear SVM models, 
where noise from sample randomness is $O(1/\sqrt{n})$ and the noise to preserve privacy is $O(1/n)$. 
The mechanism in \cite{DBLP:dblp_journals/ml/JiE13} also proves that the effect of noise brought by differential privacy is $O(1/n)$, 
while the effect from sample randomness is $O(1/\sqrt{n})$.

We should also consider the extent to which privacy is compatible with and related to the idea of generalization in machine learning. 
Intuitively, machine learning algorithms seek to generalize patterns gleaned from a training set
avoiding the effects of sample randomness.
Ideally, these algorithms should be robust to small changes in the empirical distribution of training data.
A model which fits too heavily to individual examples loses generalizability and is said to overfit. 
Perhaps the goals of differential privacy and generalization are compatible.